\documentclass[sigconf]{acmart}

\usepackage{booktabs} 

\usepackage{paralist}

\usepackage[ruled]{algorithm2e} 

\usepackage[colorinlistoftodos]{todonotes} 

\setcopyright{rightsretained}

\begin{document}
\title{Text Summarization Techniques: A Brief Survey} 
\author{Mehdi Allahyari}
\affiliation{%
  \institution{Computer Science Department\\
  University of Georgia\\
Athens, GA}
}
\email{mehdi@uga.edu}

\author{Seyedamin Pouriyeh}
\affiliation{%
  \institution{Computer Science Department\\
  University of Georgia\\
Athens, GA}
}
\email{pouriyeh@uga.edu}

\author{Mehdi Assefi}
  \affiliation{%
  \institution{Computer Science Department\\
  University of Georgia\\
Athens, GA}
}
\email{asf@uga.edu}

\author{Saeid Safaei}
\affiliation{%
  \institution{Computer Science Department\\
  University of Georgia\\
Athens, GA}
}
\email{ssa@uga.edu}

\author{Elizabeth D. Trippe}
\affiliation{%
  \institution{Institute of Bioinformatics\\
  University of Georgia\\
Athens, GA}
}
\email{edt37727@uga.edu}

\author{Juan B. Gutierrez}
\affiliation{%
  \institution{Department of Mathematics\\
  Institute of Bioinformatics\\
  University of Georgia\\
Athens, GA}
}
\email{jgutierr@uga.edu}

\author{Krys Kochut}
\affiliation{%
  \institution{Computer Science Department\\
  University of Georgia\\
Athens, GA}
}
\email{kochut@cs.uga.edu}

\renewcommand\shortauthors{Allahyari, M. et al}

\begin{abstract}
In recent years, there has been a explosion in the amount of text data from a variety of sources. This volume of text is an invaluable source of information and knowledge which needs to be effectively summarized to be useful.  In this review, the main approaches to automatic text summarization are described.  We review the different processes for summarization and describe the effectiveness and shortcomings of the different methods.
\end{abstract}

%
%

\ccsdesc[500]{Information systems~Document topic models}
\ccsdesc[500]{Information systems~Information extraction}
\ccsdesc[500]{Information systems~Summarization}

%
%

\keywords{ text summarization, knowledge bases, topic models }

\maketitle


\section{Introduction}
With the dramatic growth of the Internet, people are overwhelmed by the tremendous amount of online information and documents. This expanding availability of documents has demanded exhaustive research in the area of automatic text summarization. According to Radef et al. \cite{radev2002introduction} a \emph{summary} is defined as ``a text that is produced from one or more texts, that conveys important information in the original text(s), and that is no longer than half of the original text(s) and usually, significantly less than that''. 

\emph{Automatic text summarization} is the task of producing a concise and fluent summary while preserving key information content and overall meaning. In recent years, numerous approaches have been developed for automatic text summarization and applied widely in various domains. For example, search engines generate snippets as the previews of the documents \cite{turpin2007fast}. Other examples include news websites which produce condensed descriptions of news topics usually as headlines to facilitate browsing or knowledge extractive approaches \cite{allahyari2017brief, savova2010mayo, trippe2017vision}.

Automatic text summarization is very challenging, because when we as humans summarize a piece of text, we usually read it entirely to develop our understanding, and then write a summary highlighting its main points. Since computers lack human knowledge and language capability, it makes automatic text summarization a very difficult and non-trivial task.

Automatic text summarization gained attraction as early as the 1950s. An important research of these days was \cite{luhn1958automatic} for summarizing scientific documents. Luhn et al. \cite{luhn1958automatic} introduced a method to extract salient sentences from the text using features such as \emph{word} and \emph{phrase frequency}. They proposed to weight the sentences of a document as a function of high frequency words, ignoring very high frequency common words. Edmundson et al. \cite{edmundson1969new} described a paradigm based on \emph{key phrases} which in addition to standard frequency depending weights, used the following three methods to determine the sentence weight:

\begin{enumerate}
	\item Cue Method: The relevance of a sentence is calculated based on the presence or absence of certain cue words in the cue dictionary.
	\item Title Method: The weight of a sentence is computed as the sum of all the content words appearing in the title and headings of a text.
	\item Location Method: This method assumes that sentences appearing in the beginning of document as well as the beginning of individual paragraphs have a higher probability of being relevant.
\end{enumerate}

Since then, many works have been published to address the problem of automatic text summarization (see   \cite{gupta2010survey,erkan2004lexrank} for more information about more advanced techniques until 2000s).\\

In general, there are two different approaches for automatic summarization: \emph{extraction} and \emph{abstraction}. \emph{Extractive summarization} methods work by identifying important sections of the text and generating them verbatim; thus, they depend only on extraction of sentences from the original text. In contrast, \emph{abstractive summarization} methods aim at producing important material in a new way. In other words, they interpret and examine the text using advanced natural language techniques in order to generate a new shorter text that conveys the most critical information from the original text. Even though summaries created by humans are usually not extractive, most of the summarization research today has focused on extractive summarization. Purely extractive summaries often times give better results compared to automatic abstractive summaries \cite{erkan2004lexrank}. This is because of the fact that abstractive summarization methods cope with problems such as semantic representation, inference and natural language generation which are relatively harder than data-driven approaches such as sentence extraction. As a matter of fact, there is no completely abstractive summarization system today. Existing abstractive summarizers often rely on an extractive preprocessing component to produce the abstract of the text \cite{knight2000statistics,berg2011jointly}.

Consequently, in this paper we focus on extractive summarization methods and provide an overview of some of the most dominant approaches in this category. There are a number of papers that provide extensive overviews of text summarization techniques and systems \cite{sparck2007automatic,lloret2012text,nenkova2012survey,saggion2013automatic}.

\section{Extractive Summarization}
As mentioned before, extractive summarization techniques produce summaries by choosing a subset of the sentences in the original text. These summaries contain the most important sentences of the input. Input can be a single document or multiple documents.

In order to better understand how summarization systems work, we describe three fairly independent tasks which all summarizers perform \cite{nenkova2012survey}: 
\begin{inparaenum}[\itshape 1\upshape)]
\item Construct an intermediate representation of the input text which expresses the main aspects of the text.
\item Score the sentences based on the representation.
\item select a summary comprising of a number of sentences.
\end{inparaenum}

\subsection{Intermediate Representation}
Every summarization system creates some intermediate representation of the text it intends to summarize and finds salient content based on this representation. There are two types of approaches based on the representation: \emph{topic representation} and \emph{indicator representation}. \emph{Topic representation} approaches transform the text into an intermediate representation and interpret the topic(s) discussed in the text.
Topic representation-based summarization techniques differ in terms of their complexity and representation model, and are divided into frequency-driven approaches, topic word approaches, latent semantic analysis and Bayesian topic models \cite{nenkova2012survey}. We elaborate topic representation approaches in the following sections. \emph{Indicator representation} approaches describe every sentence as a list of features (indicators) of importance such as sentence length, position in the document, having certain phrases, etc.

\subsection{Sentence Score}
When the intermediate representation is generated, we assign an \emph{importance score} to each sentence. In topic representation approaches, the score of a sentence represents how well the sentence explains some of the most important topics of the text. In most of the indicator representation methods, the score is computed by aggregating the evidence from different indicators. Machine learning techniques are often used to find indicator weights.

\subsection{Summary Sentences Selection}
Eventually, the summarizer system selects the top $k$ most important sentences to produce a summary. Some approaches use greedy algorithms to select the important sentences and some approaches may convert the selection of sentences into an optimization problem where a collection of sentences is chosen, considering the constraint that it should maximize overall importance and coherency and minimize the redundancy. There are other factors that should be taken into consideration while selecting the important sentences. For example, context in which the summary is created may be helpful in deciding the importance. Type of the document (e.g. news article, email, scientific paper) is another factor which may impact selecting the sentences.

\section{Topic Representation Approaches} 
In this section we describe some of the most widely used topic representation approaches.

\subsection{Topic Words}
The topic words technique is one of the common topic representation approaches which aims to identify words that describe the topic of the input document. \cite{luhn1958automatic} was one the earliest works that leveraged this method by using frequency thresholds to locate the descriptive words in the document and represent the topic of the document. A more advanced version of Luhn's idea was presented in \cite{dunning1993accurate} in which they used log-likelihood ratio test to identify explanatory words which in summarization literature are called the ``topic signature''. Utilizing topic signature words as topic representation was very effective and increased the accuracy of multi-document summarization in the news domain \cite{harabagiu2005topic}. For more information about log-likelihood ratio test, see \cite{nenkova2012survey}.

There are two ways to compute the importance of a sentence: as a function of the number of topic signatures it contains, or as the proportion of the topic signatures in the sentence. Both sentence scoring functions relate to the same topic representation, however, they might assign different scores to sentences. The first method may assign higher scores to longer sentences, because they have more words. The second approach measures the density of the topic words.

\subsection{Frequency-driven Approaches}
When assigning weights of words in topic representations, we can think of binary (0 or 1) or real-value (continuous) weights and decide which words are more correlated to the topic. The two most common techniques in this category are: \emph{word probability} and \emph{TFIDF} (Term Frequency Inverse Document Frequency).

\subsubsection{Word Probability}
The simplest method to use frequency of words as indicators of importance is \emph{word probability}. The probability of a word $w$ is determined as the number of occurrences of the word, $f(w)$, divided by the number of all words in the input (which can be a single document or multiple documents):

\begin{equation}
	P(w) = \dfrac{f(w)}{N}
\end{equation}

Vanderwende et al. \cite{vanderwende2007beyond} proposed the SumBasic system which uses only the word probability approach to determine sentence importance. For each sentence, $S_j$, in the input, it assigns a weight equal to the average probability of the words in the sentence:

\begin{equation}
	g(S_j) = \dfrac{\sum_{w_i \in S_j} P(w_i)}{|\{w_i| w_i \in S_j\}|}
\end{equation}
\noindent
where $g(S_j)$ is the weight of sentence $S_j$.

In the next step, it picks the best scoring sentence that contains the highest probability word. This step ensures that the highest probability word, which represents the topic of the document at that point, is included in the summary. Then for each word in the chosen sentence, the weight is updated:

\begin{equation}
	p_{new}(w_i) = p_{old}(w_i) p_{old}(w_i)
\end{equation}

This word weight update indicates that the probability of a word appearing in the summary is lower than a word occurring once. The aforementioned selection steps will repeat until the desired length summary is reached. The sentence selection approach used by SumBasic is based on the greedy strategy. Yih et al. \cite{yih2007multi} used an optimization approach (as sentence selection strategy) to maximize the occurrence of the important words globally over the entire summary. \cite{alguliev2011mcmr} is another example of using an optimization approach.

\subsubsection{TFIDF}
Since word probability techniques depend on a stop word list in order to not consider them in the summary and because deciding which words to put in the stop list is not very straight forward, there is a need for more advanced techniques. One of the more advanced and very typical methods to give weight to words is TFIDF (Term Frequency Inverse Document Frequency). This weighting technique assesses the importance of words and identifies very common words (that should be omitted from consideration) in the document(s) by giving low weights to words appearing in most documents. The weight of each word $w$ in document $d$ is computed as follows:
\begin{equation}
	q(w) = f_d(w) * log \dfrac{|D|}{f_D(w)}
\end{equation} 

where $f_d(w)$ is term frequency of word $w$ in the document $d$, $f_D(w)$ is the number of documents that contain word $w$ and $|D|$ is the number of documents in the collection $D$. For more information about TFIDF and other term weighting schemes, see \cite{salton1988term}. TFIDF weights are easy and fast to compute and also are good measures for determining the importance of sentences, therefore many existing summarizers \cite{erkan2004lexrank,alguliev2011mcmr,alguliev2013multiple} have utilized this technique (or some form of it).

\emph{Centroid-based summarization}, another set of techniques which has become a common baseline, is based on TFIDF topic representation. This kind of method ranks sentences by computing their salience using a set of features. A complete overview of the centroid-based approach is available in \cite{radev2004centroid} but we outline briefly the basic idea. 

The first step is topic detection and documents that describe the same topic clustered together. To achieve this goal, TFIDF vector representations of the documents are created and those words whose TFIDF scores are below a threshold are removed. Then, a clustering algorithm is run over the TFIDF vectors, consecutively adding documents to clusters and recomputing the centroids according to:

\begin{equation}
	\boldsymbol{c}_j = \dfrac{\sum_{d \in C_j} d}{|C_j|}
\end{equation}

where $\boldsymbol{c}_j$ is the centroid of the $j$th cluster and $C_j$ is the set of documents that belong to that cluster. \emph{Centroids} can be considered as pseudo-documents that consist of those words whose TFIDF scores are higher than the threshold and form the cluster.

The second step is using centroids to identify sentences in each cluster that are central to topic of the entire cluster. To accomplish this goal, two metrics are defined \cite{radev2000centroid}: \emph{cluster-based relative utility} (CBRU) and \emph{cross-sentence informational subsumption} (CSIS). CBRU decides how relevant a particular sentence is to the general topic of the entire cluster and CSIS measure redundancy among sentences. In order to approximate two metrics, three features (i.e. central value, positional value and first-sentence overlap) are used.  Next, the final score of each sentence is computed and the selection of sentences is determined. For another related work, see \cite{wan2008multi}.

\subsection{Latent Semantic Analysis }
Latent semantic analysis (LSA) which is introduced by \cite{deerwester1990indexing}, is an unsupervised method for extracting a representation of text semantics based on observed words. Gong and Liu \cite{gong2001generic} initially proposed a method using LSA to select highly ranked sentences for single and multi-document summarization in the news domain. The LSA method first builds a term-sentence matrix ($n$ by $m$ matrix), where each row corresponds to a word from the input ($n$ words) and each column corresponds to a sentence ($m$ sentences). Each entry $a_{ij}$ of the matrix is the weight of the word $i$ in sentence $j$. The weights of the words are computed by TFIDF technique and if a sentence does not have a word the weight of that word in the sentence is zero. Then singular value decomposition (SVD) is used on the matrix and transforms the matrix $A$ into three matrices: $A = U\Sigma V^T$.

Matrix $U$ ($n \times m$) represents a term-topic matrix having weights of words. Matrix $\Sigma$ is a diagonal matrix ($m \times m$) where each row $i$ corresponds to the weight of a topic $i$. Matrix $V^T$ is the topic-sentence matrix. The matrix $D = \Sigma V^T$ describes how much a sentence represent a topic, thus, $d_{ij}$ shows the weight of the topic $i$ in sentence $j$.

Gong and Liu's method was to choose one sentence per each topic, therefore, based on the length of summary in terms of sentences, they retained the number of topics. This strategy has a drawback due to the fact that a topic may need more than one sentence to convey its information. Consequently, alternative solutions were proposed to improve the performance of LSA-based techniques for summarization. One enhancement was to leverage the weight of each topic to decide the relative size of the summary that should cover the topic, which gives the flexibility of having a variable number of sentences. Another advancement is described in \cite{steinberger2007two}. Steinberger et al. \cite{steinberger2007two} introduced a LSA-based method which achieves a significantly better performance than the original work. They realized that the sentences that discuss some of important topics are good candidates for summaries, thus, in order to locate those sentences they defined the weight of the sentence as follows:

Let $g$ be the "weight" function, then

\begin{equation}
	g(s_i) = \sqrt{\sum\limits_{j=1}^m d^2_{ij}}
\end{equation}

For other variations of LSA technique, see \cite{hachey2006dimensionality,ozsoy2010text}.

\subsection{Bayesian Topic Models}
Many of the existing multi-document summarization methods have two limitations \cite{wang2009multi}:
\begin{inparaenum}[\itshape 1\upshape)]
\item They consider the sentences as independent of each other, so topics embedded in the documents are disregarded.
\item Sentence scores computed by most existing approaches typically do not have very clear probabilistic interpretations, and many of the sentence scores are calculated using heuristics.
\end{inparaenum} 

Bayesian topic models are probabilistic models that uncover and represent the topics of documents. They are quite powerful and appealing, because they represent the information (i.e. topics) that are lost in other approaches. Their advantage in describing and representing topics in detail enables the development of summarizer systems which can determine the similarities and differences between documents to be used in summarization \cite{mani1999summarizing}.

Apart from enhancement of topic and document representation, topic models often utilize a distinct measure for scoring the sentence called Kullbak-Liebler (KL). The KL is a measure of difference (divergence) between two probability distributions $P$ and $Q$ \cite{kullback1951information}. In summarization where we have probability of words, the KL divergence of Q from P over the words $w$ is defined as :

\begin{equation}
	D_{KL}(P||Q) = \sum\limits_{w} P(w) \log \dfrac{P(w)}{Q(w)} 
\end{equation}

where $P(w)$ and $Q(w)$ are probabilities of $w$ in $P$ and $Q$.

KL divergence is an interesting method for scoring sentences in the summarization, because it shows the fact that good summaries are  intuitively similar to the input documents. It describes how the importance of words alters in the summary in comparison with the input, i.e. the KL divergence of a good summary and the input will be low. 

Probabilistic topic models have gained dramatic attention in recent years in various domains \cite{na2014mixture,chua2013automatic,ren2013personalized,hannon2011personalized,allahyari2015automatic,allahyari2016semantic,allahyari2016wise,allahyari2016discovering}. \emph{Latent Dirichlet allocation} (LDA) model is the state of the art unsupervised technique for extracting thematic information (topics) of a collection of documents. A complete review for LDA can be found in \cite{blei2003latent,steyvers2007probabilistic}, but the main idea is that documents are represented as a random mixture of latent topics, where each topic is a probability distribution over words.

LDA has been extensively used for multi-document summarization recently. For example, Daume et al. \cite{daume2006bayesian} proposed \textsc{BayeSum}, a Bayesian summarization model for query-focused summarization. Wang et al. \cite{wang2009multi} introduced a Bayesian sentence-based topic model for summarization which used both term-document and term-sentence associations. Their system achieved significance performance and outperformed many other summarization methods. Celikyilmaz et al. \cite{celikyilmaz2010hybrid} describe multi-document summarization as a prediction problem based on a two-phase hybrid model. First, they propose a hierarchical topic model to discover the topic structures of all sentences. Then, they compute the similarities of candidate sentences with human-provided summaries using a novel tree-based sentence scoring function. In the second step they make use of these
scores and train a regression model according the lexical and structural characteristics of the sentences, and employ the model to score sentences of new documents (unseen documents) to form a summary.

\section{Knowledge Bases and Automatic Summarization}
The goal of automatic text summarization is to create summaries that are similar to human-created summaries. However, in many cases, the soundness and readability of created summaries are not satisfactory, because the summaries do not cover all the semantically relevant aspects of data in an effective way. This is because many of the existing text summarization techniques do not consider the semantics of words. A step towards building more accurate summarization systems is to combine summarization techniques with knowledge bases (semantic-based or ontology-based summarizers).

The advent of human-generated knowledge bases and various ontologies in many different domains (e.g. Wikipedia, YAGO, DBpedia, etc) has opened further possibilities in text summarization , and reached increasing attention recently. For example, Henning et al. \cite{hennig2008ontology} present an approach to sentence extraction that maps sentences to concepts of an ontology. By considering the ontology features, they can improve the semantic representation of sentences which is beneficial in selection of sentences for summaries. They experimentally showed that ontology-based extraction of sentences outperforms baseline summarizers. Chen et al. \cite{chen2006query} introduce a user query-based text summarizer that uses the UMLS medical ontology to make a summary for medical text. Baralis et al. \cite{baralis2013multi} propose a Yago-based summarizer that leverages YAGO ontology \cite{suchanek2007yago} to identify key concepts in the documents. The concepts are evaluated and then used to select the most representative document sentences. Sankarasubramaniam et al. \cite{sankarasubramaniam2014text} introduce an approach that employs Wikipedia in conjunction with a graph-based ranking technique. First, they create a bipartite sentence-concept graph, and then use an iterative ranking algorithm for selecting summary sentences.

\section{The Impact of Context in Summarization}
Summarization systems often have additional evidence they can utilize in order to specify the most important topics of document(s). For example when summarizing blogs, there are discussions or comments coming after the blog post that are good sources of information to determine which parts of the blog are critical and interesting. In scientific paper summarization, there is a considerable amount of information such as cited papers and conference information which can be leveraged to identify important sentences in the original paper. In the following, we describe some the contexts in more details.

\subsection{Web Summarization}
Web pages contains lots of elements which cannot be summarized such as pictures. The textual information they have is often scarce, which makes applying text summarization techniques limited. Nonetheless,  we can consider the context of a web page, i.e. pieces of information extracted from content of all the pages linking to it, as additional material to improve summarization. The earliest research in this regard is \cite{amitay2000automatically} where they query web search engines and fetch the pages having links to the specified web page. Then they analyze the candidate pages and select the best sentences containing links to the web page heuristically. Delort et al. \cite{delort2003enhanced} extended and improved this approach by using an algorithm trying to select a sentence about the same topic that covers as many aspects of the web page as possible. 
For blog summarization, \cite{hu2007comments} proposed a method that  first derives representative words from comments and then selects important sentences from the blog post containing representative words. For more related works, see \cite{sharifi2013summarization,sharifi2010summarizing,hu2008comments}.

\subsection{Scientific Articles Summarization}
A useful source of information when summarizing a scientific paper (i.e. citation-based summarization) is to find other papers that cite the target paper and extract the sentences in which the references take place in order to identify the important aspects of the target paper. Mei et al. \cite{mei2008generating} propose a language model that gives a probability to each word in the citation context sentences. They  then score the importance of sentences in the original paper using the KL divergence method (i.e. finding the similarity between a sentence and the language model). For more information, see \cite{abu2011coherent,qazvinian2008scientific}

\subsection{Email Summarization}
Email has some distinct characteristics that indicates the aspects of both spoken conversation and written text. For example, summarization techniques must consider the interactive nature of the dialog as in spoken conversations. Nenkova et al. \cite{nenkova2004facilitating} presented early research in this regard, by proposing a method to generate a summary for the first two levels of the thread discussion. A thread consists of one or more conversations between two or more participants over time. They select a message from the root message and from each response to the root, considering the overlap with root context. Rambow et al. \cite{rambow2004summarizing} used a machine learning technique and included features related to the thread as well as features of the email structure such as position of the sentence in the tread, number of recipients, etc. Newman et al. \cite{newman2003summarizing} describe a system to summarize a full mailbox rather than a single thread by clustering messages into topical groups and then extracting summaries for each cluster.

\section{Indicator Representation Approaches}
Indicator representation approaches aim to model the representation of the text based on a set of features and use them to directly rank the sentences rather than representing the topics of the input text. Graph-based methods and machine learning techniques are often employed to determine the important sentences to be included in the summary.

\subsection{Graph Methods for Summarization}
Graph methods, which are influenced by PageRank algorithm \cite{mihalcea2004textrank}, represent the documents as a connected graph. Sentences form the vertices of the graph and edges between the sentences indicate how similar the two sentences are. A common technique employed to connect two vertices is to measure the similarity of two sentences and if it is greater then a threshold they are connected. The most often used method for similarity measure is cosine similarity with TFIDF weights for words. 

This graph representation results in two outcomes. First, the partitions (sub-graphs) included in the graph, create discrete topics covered in the documents. The second outcome is the identification of the important sentences in the document. Sentences that are connected to many other sentences in the partition are possibly the center of the graph and more likely to be included in the summary.

Graph-based methods can be used for single as well as multi-document summarization \cite{erkan2004lexrank}. Since they do not need language-specific linguistic processing other than sentence and word boundary detection, they can also be applied to various languages \cite{mihalcea2005language}. Nonetheless, using TFIDF weighting scheme for similarity measure has limitations, because it only preserves frequency of words and does not take the syntactic and semantic information into account. Thus, similarity measures based on syntactic and semantic information enhances the performance of the summarization system \cite{chali2008improving}. For more graph-based approaches, see \cite{nenkova2012survey}.

\subsection{Machine Learning for Summarization}
Machine learning approaches model the summarization as a classification problem. \cite{kupiec1995trainable} is an early research attempt at applying machine learning techniques for summarization. Kupiec et al. develop a classification function, \emph{naive-Bayes classifier}, to classify the sentences as summary sentences and non-summary sentences based on the features they have, given a training set of documents and their extractive summaries. The classification probabilities are learned statistically from the training data using Bayes' rule:

\begin{equation}
	P(s \in \mathcal{S} |F_1, F_2, \ldots, F_k) = \dfrac{P(F_1, F_2, \ldots, F_k) | s \in \mathcal{S}) P(s \in \mathcal{S})}{ P(F_1, F_2, \ldots, F_k)}
\end{equation}

where $s$ is a sentence from the document collection, $F_1, F_2, \ldots, F_k$ are features used in classification and $\mathcal{S}$ is the summary to be generated. Assuming the conditional independence between the features:

\begin{equation}
	P(s \in \mathcal{S} |F_1, F_2, \ldots, F_k) = \dfrac{\prod_{i=1}^k P(F_i | s \in \mathcal{S}) P(s \in \mathcal{S})}{\prod_{i=1}^k P(F_i)}
\end{equation}

The probability a sentence to belongs to the summary is the score of the sentence. The selected classifier plays the role of a sentence scoring function. Some of the frequent features used in summarization include the position of sentences in the document, sentence length, presence of uppercase words, similarity of the sentence to the document title, etc. Machine learning approaches have been widely used in summarization  by \cite{zhou2003web,wong2008extractive,ouyang2011applying}, to name a few. 

Naive Bayes, decision trees, support vector machines, Hidden Markov models and Conditional Random Fields are among the most common machine learning techniques used for summarization. One fundamental difference between classifiers is that sentences to be included in the summary have to be decided \emph{independently}. It turns out that methods explicitly assuming the dependency between sentences such as Hidden Markov model \cite{conroy2001text} and Conditional Random Fields \cite{shen2007document} often outperform other techniques.

One of the primary issues in utilizing supervised learning methods for summarization is that they need a set of training documents (labeled data) to train the classifier, which may not be always easily available. Researchers have proposed some alternatives to cope with this issue:
\begin{itemize}
	\item \textbf{Annotated corpora creation:} Creating annotated corpus for summarization greatly benefits the researchers, because more public benchmarks will be available which makes it easier to compare different summarization approaches together. It also lowers the risk of overfitting with a limited data. Ulrich et al. \cite{ulrich2008publicly} introduce a publicly available annotated email corpus and its creation process. However, creating annotated corpus is very time consuming and more critically, there is no standard agreement on choosing the sentences, and different people may select varied sentences to construct the summary.

	\item \textbf{Semi-supervised approaches:} Using a semi-supervised technique to train a classifier. In semi-supervised learning we utilize the unlabeled data in training. There is usually a small amount of labeled data along with a large amount of unlabeled data. For complete overview of semi-supervised learning, see \cite{chapelle2006semi}. Wong et al. \cite{wong2008extractive} proposed a semi-supervised method for extractive summarization. They co-trained two classifiers iteratively to exploit unlabeled data. In each iteration, the unlabeled training examples (sentences) with top scores are included in the labeled training set, and the two classifiers are trained on the new training data.
\end{itemize}

Machine learning methods have been shown to be very effective and successful in single and multi-document summarization, specifically in class specific summarization where classifiers are trained to locate particular type of information such as scientific paper summarization \cite{teufel2002summarizing,qazvinian2014generating,qazvinian2008scientific} and biographical summaries \cite{soares2011extracting,zhou2005multi,schiffman2001producing}.

\section{Evaluation}
Evaluation of a summary is a difficult task because there is no ideal summary for a document or a collection of documents and the definition of a good summary is an open question to large extent  \cite{saggion2013automatic}. It has been found that human summarizers have low agreement for evaluating and producing summaries. Additionally, prevalent use of various metrics and the lack of a standard evaluation metric has also caused summary evaluation to be difficult and challenging.

\subsection{Evaluation of Automatically Produced Summaries}
There have been several evaluation campaigns since the late 1990s in the US \cite{saggion2013automatic}. They include SUMMAC (1996-1998) \cite{mani2002summac}, DUC (the Document Understanding Conference, 2000-2007) \cite{Over:2007:DC:1284916.1285157}, and more recently TAC (the Text Analysis Conference, 2008-present) \footnote{\texttt{http://www.nist.gov/tac/about/index.html}}. These conferences have primary role in design of evaluation standards and evaluate the summaries based on human as well as automatic scoring of the summaries.

In order to be able to do automatic summary evaluation, we need to conquer three major difficulties: 
\begin{inparaenum}[\itshape i\upshape)]
\item It is fundamental to decide and specify the most important parts of the original text to preserve.
\item Evaluators have to automatically identify these pieces of important information in the candidate summary, since this information can be represented using disparate expressions.
\item the readability of the summary in terms of grammaticality and coherence has to be evaluated.
\end{inparaenum}

\subsection{Human Evaluation}
The simplest way to evaluate a summary is to have a human assess its quality. For example, in DUC, the judges would evaluate the coverage of the summary, i.e. how much the candidate summary covered the original given input. In more recent paradigms, in particular TAC, query-based summaries have been created. Then judges evaluate to what extent a summary answers the given query. The factors that human experts must consider when giving scores to each candidate summary are grammaticality, non redundancy, integration of most important pieces of information, structure and coherence. For more information, see \cite{saggion2013automatic}.

\subsection{Automatic Evaluation Methods}
There has been a set of metrics to automatically evaluate summaries since the early 2000s. ROUGE is the most widely used metric for automatic evaluation.

\subsubsection{ROUGE}
Lin \cite{lin2004rouge} introduced a set of metrics called Recall-Oriented Understudy for Gisting Evaluation (ROUGE) to automatically determine the quality of a summary by comparing it to human (reference) summaries. There are several variations of ROUGE (see \cite{lin2004rouge}), and here we just mention the most broadly used ones:
\begin{itemize}
	\item \textbf{ROUGE-$n$:} This metric is recall-based measure and based on comparison of $n$-grams. a series of $n$-grams (mostly two and three and rarely four) is elicited from the reference summaries and the candidate summary (automatically generated summary). Let $p$ be "the number of common $n$-grams between candidate and reference summary", and $q$ be "the number of $n$-grams extracted from the reference summary only". The score is computed as:
	\begin{equation}
		\text{ROUGE-}n = \dfrac{p}{q}
	\end{equation}

	\item \textbf{ROUGE-$L$:} This measure employs the concept of \emph{longest common subsequence} (LCS) between the two sequences of text. The intuition is that the longer the LCS between two summary sentences, the more similar they are. Although this metric is more flexible than the previous one, it has a drawback that all $n$-grams must be consecutive. For more information about this metric and its refined metric, see \cite{lin2004rouge}.

	\item \textbf{ROUGE-SU:} This metric called \emph{skip bi-gram and uni-gram} ROUGE and considers bi-grams as well as uni-grams. This metric allows insertion of words between the first and the last words of the bi-grams, so they do not need to be consecutive sequences of words. 
\end{itemize}

\section{Conclusions}
The increasing growth of the Internet has made a huge amount of information available. It is difficult for humans to summarize large amounts of text. Thus, there is an immense need for automatic summarization tools in this age of information overload.

In this paper, we emphasized various extractive approaches for single and multi-document summarization. We described some of the most extensively used methods such as topic representation approaches, frequency-driven methods, graph-based and machine learning techniques. Although it is not feasible to explain all diverse algorithms and approaches comprehensively in this paper, we think it provides a good insight into recent trends and progresses in automatic summarization methods and describes the state-of-the-art in this research area.

\begin{acks}
This project was funded in part by Federal funds from the US National Institute of Allergy and Infectious Diseases, National Institutes of Health, Department of Health and Human Services under contract \#HHSN272201200031C, which supports the Malaria Host-Pathogen Interaction Center (MaHPIC).
\end{acks}

\section{Conflict of Interest}
The author(s) declare(s) that there is no conflict of interest regarding the publication of this article.

\bibliographystyle{ACM-Reference-Format}
\bibliography{compbib}

\end{document}